\documentclass[preprint,5p,times,twocolumn]{elsarticle}

\usepackage{booktabs} 
\usepackage{multirow}
\usepackage{multicol}
\usepackage{amsmath}
\usepackage{bbm}
\usepackage[ruled,vlined]{algorithm2e}
\usepackage{color}

\SetKwInOut{Parameter}{Parameter}

\usepackage{amssymb}

\usepackage{lineno}

\journal{Pattern Recognition}

\begin{document}

\begin{frontmatter}



\title{Unsupervised Representation Learning by Discovering Reliable Image Relations}


\author[1]{Timo Milbich \corref{cor1} \fnref{f1}}
\fntext[f1]{Authors contributed equally to this work.}
\fntext[f2]{Mail: https://hci.iwr.uni-heidelberg.de/compvis.}

\author[2]{Omair Ghori \fnref{f1}}

\author[3]{Ferran Diego}

\author[1]{Bj\"orn Ommer}

\address[1]{Heidelberg Collaboratory for Image Processing and Interdisciplinary Center for Scientific Computing (HCI), Heidelberg University, Germany}
\address[2]{Robert Bosch GmbH, Hildesheim, Germany}
\address[3]{Telefonica R\&D, Barcelona, Spain}

\begin{abstract}
   Learning robust representations that allow to reliably establish relations between images is of paramount importance for virtually all of computer vision. Annotating the quadratic number of pairwise relations between training images is simply not feasible, while unsupervised inference is prone to noise, thus leaving the vast majority of these relations to be unreliable. To nevertheless find those relations which can be reliably utilized for learning, we follow a divide-and-conquer strategy: We find reliable similarities by extracting compact groups of images and reliable dissimilarities by partitioning these groups into subsets, converting the complicated overall problem into few reliable local subproblems. For each of the subsets we obtain a representation by learning a mapping to a target feature space so that their reliable relations are kept. Transitivity relations between the subsets are then exploited to consolidate the local solutions into a concerted global representation. While iterating between grouping, partitioning, and learning, we can successively use more and more reliable relations which, in turn, improves our image representation. In experiments, our approach shows state-of-the-art performance on unsupervised classification on ImageNet with $46.0\%$ and competes favorably on different transfer learning tasks on PASCAL VOC. 

\end{abstract}


\begin{keyword}
Unsupervised Learning \sep Visual Representation Learning \sep Unsupervised Image Classification \sep Mining Reliable Relations \sep Divide-and-Conquer

\vspace{3mm}
© 2019. This manuscript version is made available under the CC-BY-NC-ND 4.0 license http://creativecommons.org/licenses/by-nc-nd/4.0/



\end{keyword}

\end{frontmatter}


\section{Introduction}
The driving force of deep learning has been supervised training using vast amounts of tediously labeled training samples, such as object bounding boxes for visual recognition. Since easily accessible visual data is growing exponentially, manual labeling of training samples constitutes a bottleneck to utilizing all this valuable data. Consequently, there has recently been great interest in weakly supervised \cite{weakly_sup}, self-supervised \cite{norooziECCV16}, and unsupervised \cite{Xiong2017, Milbich_2017_ICCV} approaches to representation learning. Fundamental computer vision problems like classification~\cite{class_pr, rubio_PR_2015}, object detection~\cite{hyperfusionnet} and image segmentation~\cite{feat_for_seg} all directly depend on such learned representations to find similar objects or group related image areas.
\\
To learn a characteristic representation of images and the distances between them, different degrees of supervision can be considered: \emph{(i)} supervised learning using samples with class labels \cite{sup_hashing}, \emph{(ii)} user feedback providing weakly supervised side information in terms of pairwise constraints \cite{npair}, \emph{(iii)} problem specific surrogate tasks such as colorization \cite{larsson2017colorproxy}, permutations \cite{norooziECCV16}, or transitivity \cite{Wang_2017_ICCV}, and \emph{(iv)} unsupervised feature learning \cite{pmlr-v70-bojanowski17a}. Regardless of the training signal, be it unaries such as class labels \cite{cliqueCNN}, binary similarity constraints between samples \cite{npair} or sample ordering constraints \cite{fuzzy_dml}, a dataset of $N$ training samples gives rise to $N^2$ pairwise relations, exploitable for learning our representation. In the absence of supervisory information, these relations need to be automatically inferred during training. However, the vast majority of these inferred pairwise relations turn out to be unreliable as discussed in Sect. \ref{sec:app_intro}, Fig. \ref{fig:cliquReliability}, and Fig. \ref{fig:sigmoidDist}. Despite the danger of diminished performance due to learning from spurious relations, recent approaches on unsupervised representation learning~\cite{caron2018deep,pmlr-v70-bojanowski17a}, nevertheless, do not question the reliability of these relations. Now, assuming that only a small fraction of correct relations per sample can be identified reliably (i.e. we are left with at most $O(N)$ class labels or pairwise link constraints), how can we discover those few reliable relations, when no label or guiding side information is available?

\begin{figure*}[t]
  \centering
  \includegraphics[width=16cm]{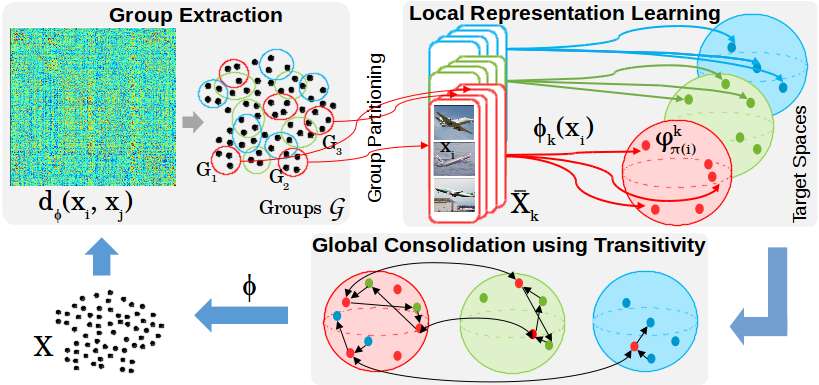}
\caption{
Overview of our iterative learning procedure. We first find reliable similarity constraints by forming compact groups. To avoid unreliable dissimilarities, we partition the data into sets of mutually dissimilar groups, $\bar{X}_k$. Based on these (dis-)similarity constraints between/within groups, we learn a local representation $\phi_k(x_i)$ for subset $\bar{X}_k$. Finally, we exploit sparse couplings between the local representations to arrive at a consolidated global representation. This iterative procedure improves the overall representation by successively adding reliable constraints into the learning process.
} 
\label{fig:pipeline}
\end{figure*}

In this work we propose a novel approach to visual representation learning that explicitly identifies and leverages reliable image relations without the need for annotations, supervision, problem-specific surrogate tasks for self-supervision, or pre-training. By extracting compact groups of images we are able to harness reliable similarities. Subsequently, we divide these compact groups constituting the overall learning problem into smaller, (potentially overlapping) subproblems, such that each contains only reliable dissimilarities between their groups. Thus, whereas the complicated global problem suffers from many of the $N^2$ relations not being reliable, we ensure that the samples in each subproblem are either reliably similar or dissimilar. Optimization is then performed by learning a mapping from the images into a dedicated target space, built to reflect the structure and distribution estimated from the reliable relations for each subset. Next, coupling the local subproblems by utilizing transitivity between their samples allows us to consolidate the learned individual representations into a concerted global representation. Finally, by alternating between extracting reliable relations and learning, we successively incorporate more reliable relations and in turn more data which ultimately improves our image representation (cf. Fig. \ref{fig:pipeline}).
\\
We evaluate our model on challenging benchmarks and achiev state-of-the-art performance on the ImageNet dataset, thus proving the scalability of our approach. Further, our approach performs comparably to the state-of-the-art in transfer learning on PASCAL VOC indicating its general applicability. By performing ablation and analysis studies, we finally provide insights into our learning procedure.


\section{Related Work}
Leveraging explicit relationships between training images for representation learning is widely studied using different orders of constraints (e.g. binary similarities \cite{npair}, ranking constraints \cite{fuzzy_dml}). All these methods use label information or strong pretraining as the performance of such task heavily depends on the quality of the pairwise constraints. However, densely labeling all pairwise constraints is infeasible.
\\
Due to the difficulty of extracting reliable relations from data, many recent label-free approaches resort to generic prior assumptions on the data distribution or single image- and class-based tasks. (Deep) clustering methods \cite{caron2018deep, cliqueCNN} rely on a predefined number of pseudo-classes which typically is estimated by heuristics and further focus on similarity constraints only. Our model explicitly models both similarity and dissimilarity constraints estimated from data itself. Bojanowski et al.\cite{pmlr-v70-bojanowski17a} find a mapping between images and a uniformly discretized target space, thus enforcing their representation to resemble a distribution of pairwise relationships independent of the actual data structure. Sanakoyeu et al. \cite{cliqueCNN} cluster data into small surrogate classes to perform a global classification task, however, not considering that similar images may end up in competing, different classes. Thus, inferred relationships during training suffer from contradicting training signals. Also DeepCluster \cite{caron2018deep} follows this strategy based on disjoint k-means clustering, thus enforcing clear distinct boundaries which potentially disagree with the real data distribution. In contrast, our grouping process does not enforce hard class boundaries and is able to adapt to the data structure. Moreover by splitting groups into reliable subproblems and constructing a learning problem following their distance distribution, our groups corroborate their training signal. Dosovitski et al. \cite{exemplarCNN} cast distance learning as an exemplar classification task utilizing heavy data augmentations at the cost of poor scalability to large data collections.
\\
Self--supervised learning approaches aim to leverage data itself by typically solving surrogate tasks based on temporal \cite{Wang_2015_ICCV} and spatial \cite{norooziECCV16} coherence. 
These approaches are either domain specific or operate on images independently thus missing out on their relationships. Our work, in contrast, explicitly models relationships between images. Gidaris et al. \cite{gidaris2018unsupervised} exploit image geometry and classify rotations applied to input images. Even though they report good results on image classification tasks on large datasets, this task is conceptually dependent on large variations in the underlying data distribution to avoid trivial image representation, potentially missing out fine-grained relationships between images.
\\
Generative models based on GAN\cite{gan}- and VAE\cite{vae}-like architectures recently became a popular choice for unsupervised learning. These approaches learn mappings between images and a latent space driven by a generative task and thus implicitly learn an image representation. Unfortunately such approaches typically suffer from limited applicability for large datasets with high variance due to the difficult optimization problem. Further their training is regularized using data-independent priors, e.g. by enforcing the learned feature space to follow a gaussian distribution. Our approach on the other side explicitly learns an image representation using data dependent constraints and scales to large datasets.

\section{Approach}
\label{sec:app_intro}

\begin{figure}[t] 
    \centering
    \includegraphics[width=0.9\columnwidth]{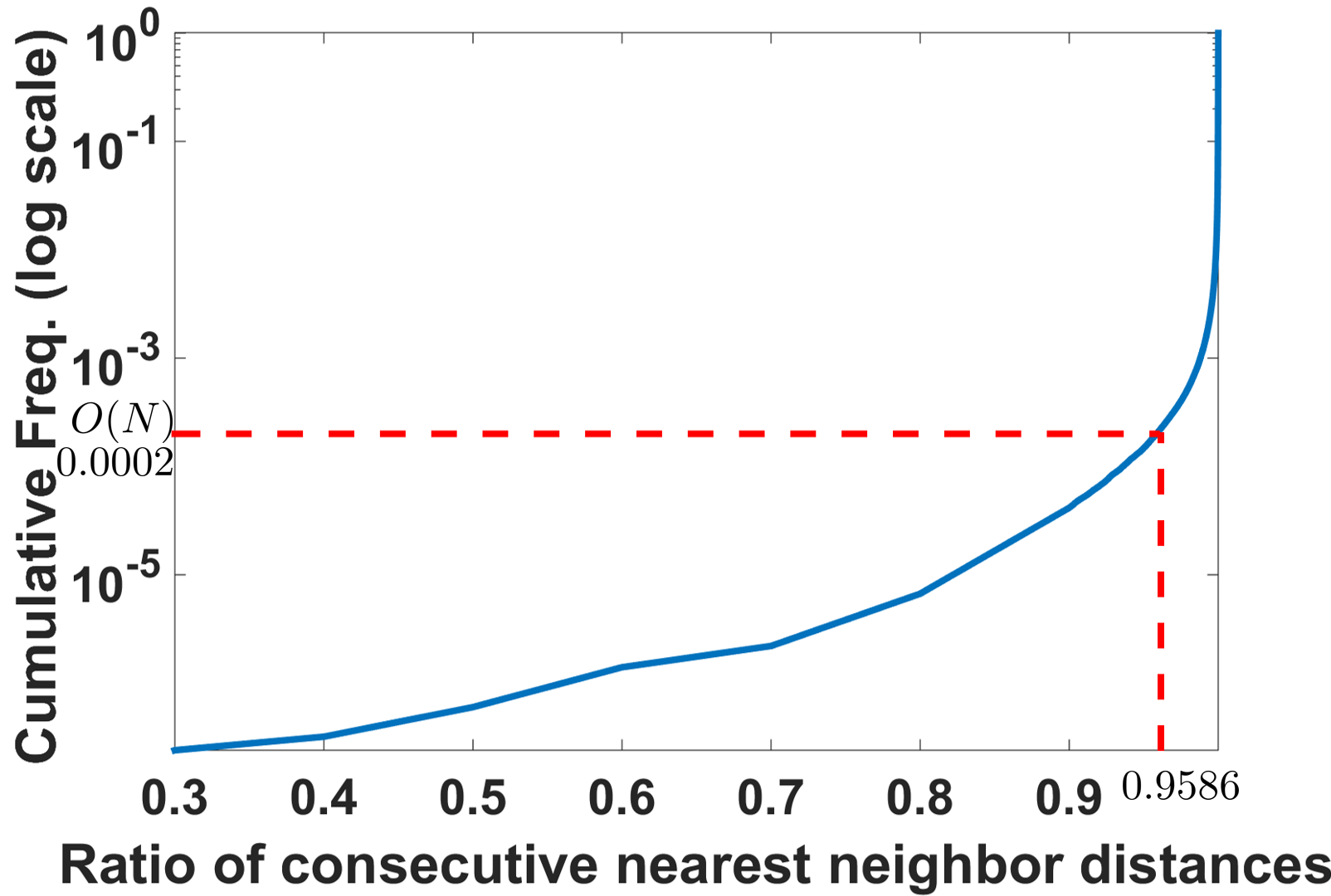}
    \caption{
    Nearest neighbour distance ratios. For most of the $N^2$ pairwise relations, the ratio $^{d(x_i,x_j)}/_{d(x_i,x_{j+1})}$ (for sorted neighbors $x_j$) is close to 1. Only $O(N)$ have a robust ordering. This analysis is based on $N=5000$ samples from the STL-10\cite{stl10} dataset using euclidean distances based on an unsupervised representation.
    }
    \label{fig:cliquReliability}
\end{figure}

Let us now learn a representation $\phi: X \rightarrow \Phi$ that allows to relate image samples $x_i, x_j \in X$ to another. This is equivalent to learning a distance $d_{\phi}\left( \phi(x_i), \phi(x_j)  \right) \stackrel{!}{=} d(x_i, x_j)$, i.e. learning a representation $\phi$ such that given image relations $d(x_i, x_j)$ are reflected and preserved in the embedding space $\Phi$. Thus, learning $\phi$ is propelled by pairwise relationships between images indicated by $d(x_i,x_j)$: In supervised training $d(x_i, x_j)$ is typically defined on the basis of manually provided class labels $l(x_i)$, weak user feedback, problem specific surrogate tasks, dense triplet ranking constraints such as $d(x_i,x_j) < d(x_i,x_k)$, or other sparse partial ordering constraints.

\subsection{Reliable Relations for Learning a Representation:} 
\begin{figure}[t]
\centering
  \centering
  \includegraphics[width=0.85\columnwidth]{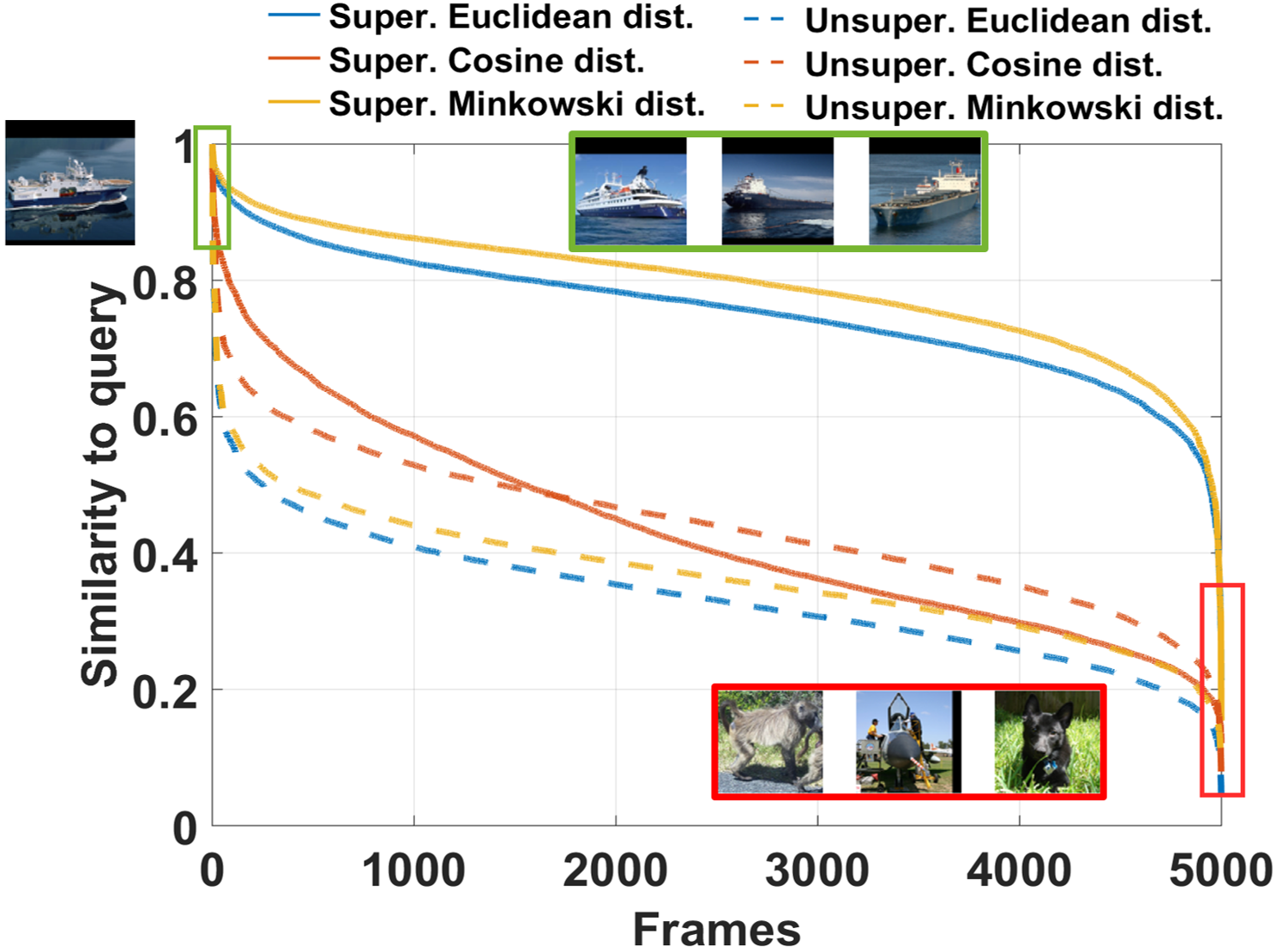} 
\caption{
Sorted pairwise similarities for a query image based on different representations and distance metrics resulting from supervised and unsupervised training on STL-10\cite{stl10}. Only the strong (dis-)similarities at both ends are reliable to provide a robust ordering.
}
\label{fig:sigmoidDist}
\end{figure}

Regardless of the origin of $d(x_i, x_j)$, only a small number of all possible $N^2$ pairwise relations (for $N$ training samples) may be feasibly provided by manual annotation. For the particular case of unsupervised learning only a small number of the pairwise relations can be inferred correctly for training with high confidence. Let's consider a triplet of images $x_i, x_j, x_k$ with the ground-truth distance between $x_i$ and $x_j$ being small and between $x_i$ and $x_k$ being large. A learned distance is correct, if it obeys these ground-truth constraints. Furthermore to ensure robustness to noise these constraints should be obeyed reliably by a clear margin $d(x_i,x_j) / d(x_i,x_k) \overset{!}{\ll} 1$. In Fig. \ref{fig:cliquReliability} we plot this ratio for consecutive nearest neighbors. This is the same ratio used in \cite{lowe:sift} to measure the reliability of a matching similar image. We observe that for only $O(N)$ pairwise distances the ratio is significantly smaller than $0.95$. All other relations would not even have an opportunity to exhibit above correctness constraints reliably and, thus, would be falsely identified to be correct rather than spurious.
Further, in Fig. \ref{fig:sigmoidDist} we plot the sorted similarities for a query image. Observe from Fig. \ref{fig:cliquReliability} that above triplet constraints can only be fulfilled where there is significant slope in Fig. \ref{fig:sigmoidDist}. Thus, only relations from both ends, where we have strong (dis-)similarities, can be considered to be reliable. 
These relations are significantly less susceptible to change under noise than the vast majority, as analyzed in Fig. \ref{fig:sens_to_noise}. However, recent work on unsupervised learning has nevertheless simply relied on all pairwise relations \cite{pmlr-v70-bojanowski17a} inferred during training at the cost of incorporating corrupted relations. In contrast, we now present an approach for unsupervised representation learning which explicitly aims at extracting and leveraging these reliable relations.

\noindent
\subsection{Outline of our Iterative Representation Learning:} We first decompose the training set into subsets $\bar{X}_k \subset X, k = 1,\dots,K$ of images by extracting (Sec. \ref{sec:reliableGroups}) and dividing (Sec. \ref{sec:partitioning}) potentially overlapping compact groups of images, exhibiting reliable mutual similarity, based on the representation from the previous training iteration. Within each $\bar{X}_k$ all mutual image relations are reliable. In each iteration, learning a representation $\phi$ (Sec. \ref{sec:learning_all}) then proceeds as follows: For each subset $\bar{X}_k$ we seek an embedding $\phi_k: x_i\in \bar{X}_k \mapsto \phi_k(x_i) \in \Phi_k$. To learn $\phi_k$, we randomly sample target points $\varphi_t^k$ such that the distribution of their pairwise distances matches those of the $x_i\in \bar{X}_k$. Learning the mapping from images $x_i$ to targets $\varphi_t^k$ then yields the local representations $\phi_k$ (Sec. \ref{sec:local_learning}). In a final step, all these local representations are merged into a single overall representation $\phi$ by exploiting transitivity relations between those samples $x_i$ which are shared among subsets $\bar{X}_k$ (Sec. \ref{sec:coupling}). Based on representation $\phi$, reliable relations are then again extracted to serve as input for the next iteration. Since no annotations are provided, the first iteration of our training starts \emph{from scratch} with a randomly initialized CNN and random assignments.

\subsection{Compact Groups for Finding Reliable Similarities}
\label{sec:reliableGroups}

\begin{figure}[t]
\centering
\includegraphics[width=0.95\columnwidth]{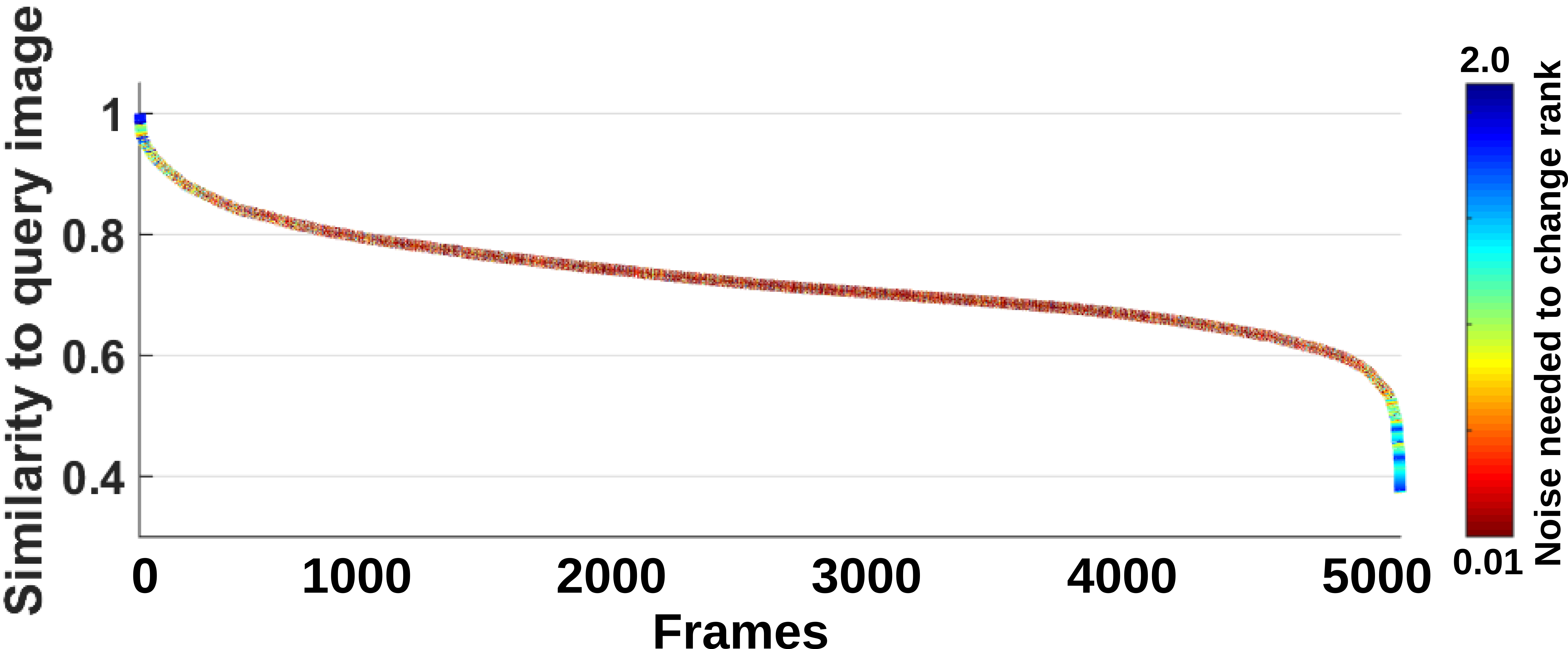} 
\caption{
Sorted pairwise image similarities for an STL-10 query image. Color indicates amount of noise (gaussian, $\mu = 0$, $\sigma^2 = 0.01,\dots,2$) needed for each image to change its rank w.r.t. query.
}
\label{fig:sens_to_noise}
\end{figure}

\begin{figure}[t]
\centering
\begin{tabular}{cc}
  \centering
      \includegraphics[width=0.47\columnwidth]{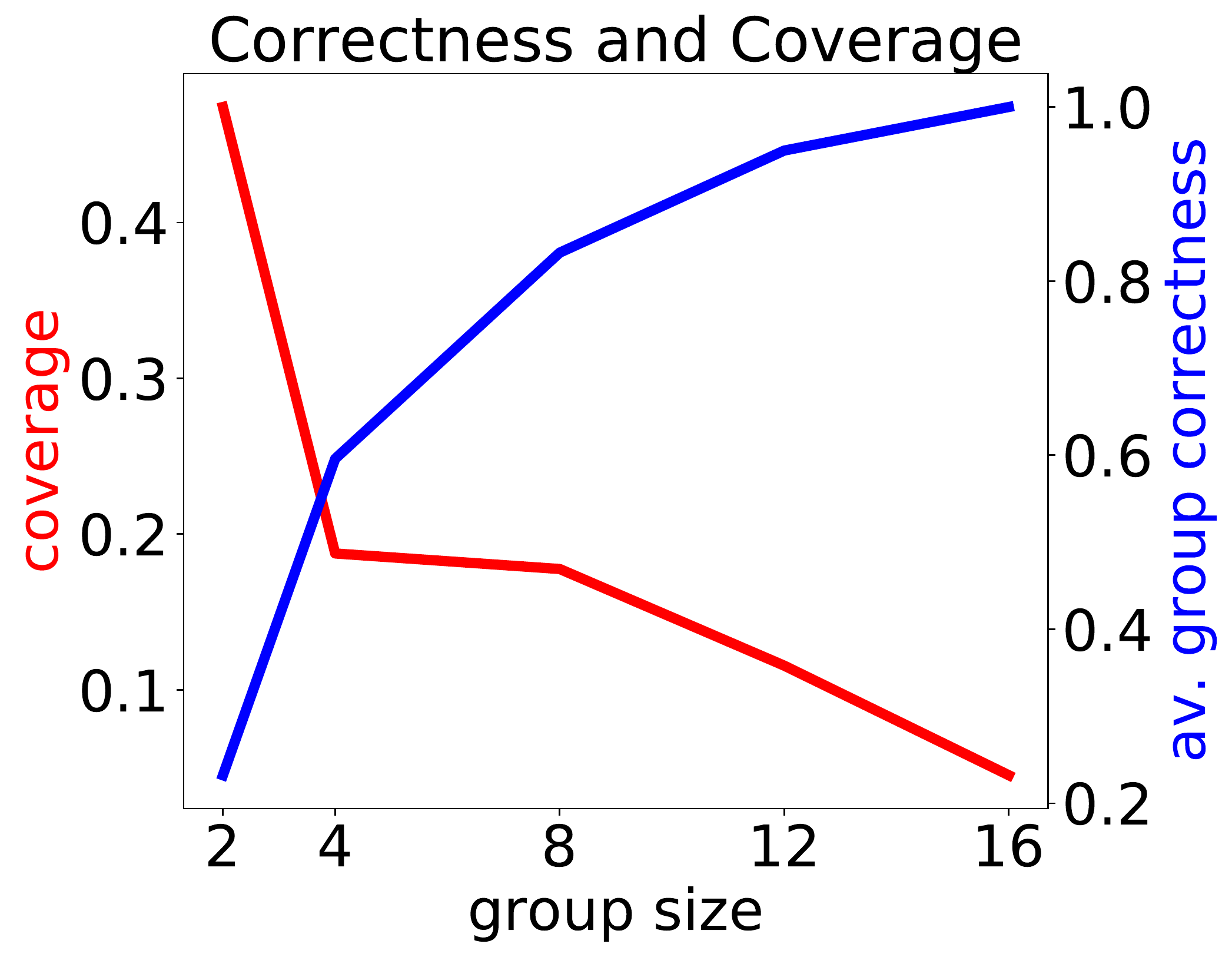} & 
  \includegraphics[width=0.54\columnwidth]{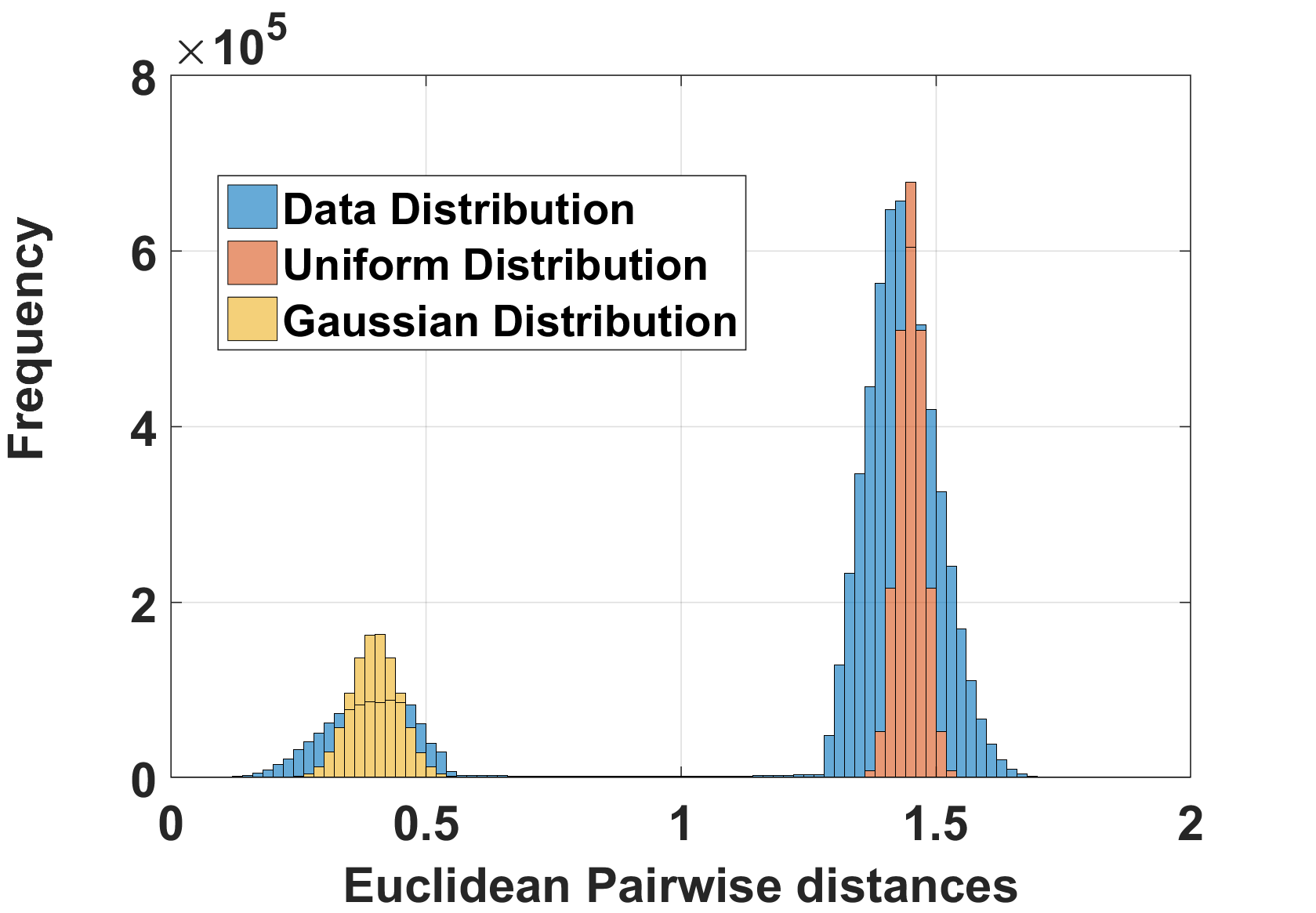} \\
    (a) & (b)
\end{tabular}
\caption{a) Group correctness vs. data coverage w.r.t group size. blue: average correctness of relations in a group. red: data coverage for groups $G$ of different size extracted on ImageNet using our representation. We call a relation correct if it links images with the same label. Note that we only use labels for evaluation in this plot, but not for extracting our groups. Coverage: fraction of all samples that can be covered by compact groups of a given size relative to the overall number of extracted groups $G$. 
b) Data- vs. Target-Distribution. Distribution of all pairwise intra-group and inter-group distances (blue) based on a fully supervised trained representation, of points uniformly sampled from an $\ell_2$ unit sphere (orange), and of a Gaussian distribution (yellow). Evidently, most data points are far apart, approximated by the orange mode. But there are also characteristic compact hubs approximated by the yellow mode (which has been magnified for the purpose of this illustration). Note that the sampled distributions can approximate the data distribution.
}
\label{fig:purity_over_groups}
\end{figure}

Every training iteration builds upon the distances learned in the previous round. Since the majority of inferred relations between our training samples is not reliable, how can we find the ones we can rely upon without annotations? In Fig. \ref{fig:cliquReliability} to Fig. \ref{fig:sens_to_noise} we empirically demonstrate that only a few nearest neighbours (NN) can robustly be identified. Unfortunately, even these do not always give rise to correct relations. For instance, two samples may spuriously have a small distance or a sample may be an outlier, thus corrupting the nearest neighbors for a given sample $x_i$. This issue can however be alleviated by forming compact groups of images. Fig. \ref{fig:purity_over_groups} (a) shows that considering dense groups of $h$ samples increases the chance of inferring correct similarity relations, following the intuition that in dense areas of our feature space pairwise relations do not arise accidentally but due to actual commonalities. For different group sizes $h$ on ImageNet, the plot illustrates the average percentage of correct image relations in a group based on their ground-truth labels (blue). As we observe, the chance that $h$ samples appear erroneously close to another (forming a compact group) becomes increasingly unlikely as $h$ increases. On the other hand, large compact groups are scarce and therefore only a small number of samples can be covered for large $h$ (red).
\\
Let the largest pairwise distance, $\nu_G$, between members of a group $G$ represent its compactness. Further, when building groups of random samples, it is highly unlikely to form a group with low compactness, i.e. correctly mutually close samples. Thus, we demand $\nu_G$ for each $G$ to be smaller than the $p$-th percentile of the compactness of a set of randomly built groups of equal size. Consequently, we extract reliable groups $G$ by starting with a seed $x_i$ and add its nearest neighbors as long as the resulting compactness $\nu_{G}$ is below the $p$-th percentile compactness of the randomly sampled groups of equal size. Thus, we grow $G$ to be as large as possible. We denote $\mathcal{G}$ as the resulting set of all possible (potentially overlapping) groups $G$ with reliable pairwise similarity relationships.

\subsection{Reliable Dissimilarity by Partitioning Groups }
\label{sec:partitioning}

\paragraph{Finding Reliable Dissimilarities:} The compact groups in $\mathcal{G}$ provide small distances that we can reliably use for learning a representation. However, the relationships \textit{between} the groups can be arbitrary and many are unlikely to be reliable as discussed above and in Fig. \ref{fig:cliquReliability}. Therefore, to further increase reliability, instead of using all relations in $\mathcal{G}\subset X$ for learning our representation, we partition them into subsets of groups $\bar{X}_k:= \{G_{k_1}, G_{k_2}, \ldots\}$. By distributing overlapping groups across different $\bar{X}_k$ while maximizing the distance between groups within a subset, we gathers the reliable dissimilarity relationships from the tail of the distance distributions shown in Fig. \ref{fig:sigmoidDist}. Thus, all relations in each subset are as reliable as possible.

\paragraph{Optimization Problem:} Formally, we partition $\mathcal{G}$ using the following criteria: \emph{(i)} all groups $G \in \bar{X}_k$ should be mutually distant, \emph{(ii)} partially overlapping groups should be distributed across different $\bar{X}_k$ to establish couplings between the subproblems exploitable for transitivity relations, and \emph{(iii)} the union of all subsets, $\bigcup_k \bar{X}_k$, should cover $X$ as much as possible to maximize the usage of training samples. Using these constraints, we formulate the partitioning process as the following optimization problem. 
\\
Let $\mathbf{C} \in \{0,1\}^{|\mathcal{G}| \times |X|}$ be the assignment matrix of samples $x_i$ to groups $G$. Furthermore, the column vector $a_k \in \{0,1\}^{|\mathcal{G}| \times 1}$ indicates groups assigned to subset $\bar{X}_k$. Then $A=(a_1,a_2,\dots)$ are the assignments of  groups to all $\bar{X}_k$. Moreover, $S \in \mathbb{R}^{|\mathcal{G}| \times |\mathcal{G}|}$ contains the mean pairwise distances between any two groups, $S_{kl}= ^1\!\!/_{\!|G_k \times G_l|} \sum_{(x_i,x_j) \in G_k \times G_l} d(x_i,x_j)$. 
\\
\\
The objective then becomes 

\begin{equation}
\begin{aligned}
\min_{\textbf{A}} \quad & \text{tr}(\textbf{A}^\top\textbf{S}\textbf{A} )-\text{tr}(\textbf{A}^\top\text{diag}(\textbf{S})\textbf{A}) \\
& - \lambda_1\sum\limits_{k}||a_k^\top\textbf{C}||_p^p - \lambda_2||\mathbbm{1} \textbf{A}^\top\textbf{C}||_p^p \\
\text{s.t.} \quad & \textbf{A}^\top\mathbbm{1}^\top = |\bar{X}_k| \mathbbm{1}^\top
\end{aligned}
\end{equation}

where $\text{tr}(\textbf{A}^\top\textbf{S}\textbf{A} )$ is regularized for the diagonal elements of $S$ and enforces constraint \emph{(i)}, i.e. maximal distance between groups in a $\bar{X}_k$, 
$||a_k^\top \textbf{C}||_p^p$ maximizes \emph{(ii)}, i.e. the distribution of groups across the subsets $\bar{X}_k$, 
and $||\mathbbm{1} \textbf{A}^\top\textbf{C}||_p^p$ enforces \emph{(iii)}, i.e. maximal coverage of $\mathcal{G}$. $\lambda_1, \lambda_2$ are weighting terms for adjusting relative impact of individual constraints. This optimization problem can be efficiently solved following CliqueCNN~\cite{cliqueCNN}. This work has addressed a similar problem to find a discrete partitioning of images into surrogate classes for optimizing a standard classification task equal to DeepCluster~\cite{caron2018deep}. However, both \cite{cliqueCNN} and \cite{caron2018deep} are using all resulting surrogate classes without considering their reliability. Thus, they are inevitably prone to introducing noise into the learning process. In contrast, we seek a partitioning of the set of already extracted groups $\mathcal{G}$ into subsets to further increase the reliability of our relations which we then use to construct a dedicated learning problem.

\subsection{Unsupervised Representation Learning}
\label{sec:learning_all}

\begin{figure}[t]
\begin{minipage}[b]{\columnwidth}
  \centering
  \includegraphics[width=0.99\columnwidth]{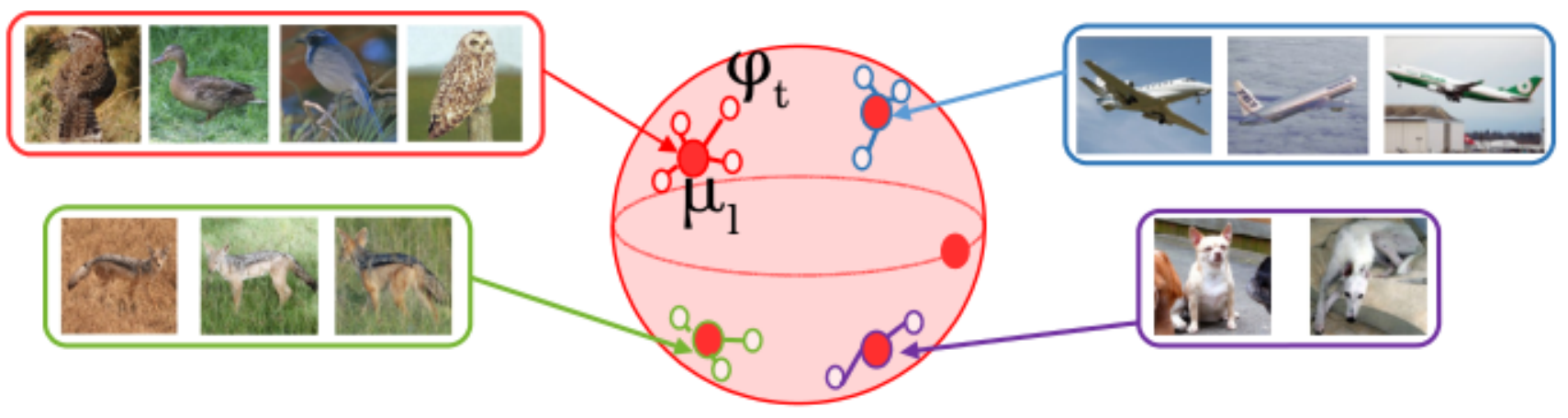}
\caption{
Learning feature vectors by sampling a target space. Large distances between groups $G_l$ are captured by sampling their centroids $\mu_l$ uniformly from the surface of a hypersphere. Then target feature points $\varphi_t$ for the $x_i \in G_l$ are sampled from a Gaussian around $\mu_l$ to represent the compact group.
}
\label{fig:cliques_on_sphere}
\end{minipage}
\end{figure}

Now we learn a representation $\phi_k(x_i)$ that preserves all the reliable relations in $\bar{X}_k$. To this end, we build a target space by randomly sampling target points $\varphi^k_t$, such that the distribution of their pairwise distances matches the distances in $\bar{X}_k$. Finding a representation for the data points $x_i$ then corresponds to solving an assignment problem $t=\pi(i)$ to find their targets $t$ while simultaneously learning the corresponding mappings.

\subsubsection{Constructing the target spaces}
 We sample the targets $\varphi^k_t$ from the surface of a high-dimensional sphere with local Gaussian hubs accounting for the compact groups $G$. More precisely, we first sample centroids $\mu_l, l=1:|\bar{X}_k|$ uniformly from the surface of the $D$-dimensional unit hypersphere (where D is the dimensionality of $\Phi_k$). Then we sample $\varphi^k_t$ from the Gaussians $\mathcal{N}(\mu_l, \Sigma)$ with fixed covariance matrix $\Sigma$ as illustrated in Fig. \ref{fig:cliques_on_sphere}. Note that this sampling process is derived from and \textit{approximates the actual data distribution} as illustrated in Fig. \ref{fig:purity_over_groups} (b). Using the representation of a fully supervised model as a proxy for ground-truth image relations (for this experiment only), we observe that the distribution of pairwise intra-group and inter-group distances (blue distribution) exhibits two distinctive modes as intuitively expected: A large mode representing mediocre to large inter-group distances and a minor second mode of small intra-group distances reflecting dense neighbourhoods of mutually similar samples. Sampling $\varphi^k_t$ based on Gaussians $\mathcal{N}(\mu_l, \Sigma)$ uniformly distributed on a hypersphere explicitly approximates this distribution (orange and yellow distributions). In contrast, other approaches such as \cite{pmlr-v70-bojanowski17a} rely on a data independent prior which does not sufficiently account for dense neighbourhoods of highly similar datapoints and consequently diminishes the expressiveness of the representation to be learned.

\subsubsection{Learning the Representations $\phi_k$}
\label{sec:local_learning}
Learning $\phi_k$ is now formulated as establishing correspondences $t=\pi(i)$ between the $x_i$ from a group $G \in \bar{X}_k$ and the targets $\varphi^k_t$.
This requires to minimize the distances between $\phi_k(x_i)$ and $\varphi^k_{\pi(i)}$. We obtain $\phi_k$ and $\pi$ by jointly minimizing the optimization problem

\begin{equation}
\mathcal{L}^k_{\text{local}} := \sum_{x_i \in G:\; G \in \bar{X}_k}
d_{\phi_k}\left(\phi_k(x_i), \varphi^k_{\pi(i)}\right) .
\end{equation}

We solve this problem by alternating between two steps: \emph{(i)} Find optimal assignments $\pi$ based on the current representation $\phi_k$. Since global solvers for such an assignment problem typically exhibit the prohibitive cost of $O(N^3)$ in the number of input samples, we adopt the efficient algorithm of \cite{pmlr-v70-bojanowski17a} which uses stochastic local updates to approximate the Hungarian method. \emph{(ii)} Given an assignment, we optimize the representation $\phi_k$ by learning a CNN with weights $\theta$ to minimize the distances $d_{\phi_k}(\phi_k(x_i), \varphi^k_{\pi(i)})$. By alternating between these two steps (we re-assign between $x_i$ and $\varphi^k_t$ every 3 epochs), the model needs to reason about which targets apply for which images and which images should be assigned next to each other. Thus it needs to find an optimal arrangement of groups $G$ on the target space preserving their relations and further needs to infer meaningful relations between groups $G$. At initialization, we start with a randomly initialized CNN and, thus, random initial weights $\theta$ and random assignments $\pi(i)$.
\\
Using this learning process, we obtain a representation $\phi_k$ for each of our subsets $\bar{X}_k$, each having its own distances $d_k(x_i, x_j)$ = $d_{\phi_k}(\phi_k(x_i), \phi_k(x_j))$, with $d_{\phi_k}$ being the $L^2$ metric in our implementation.

\subsubsection{Coupling Subproblems to Consolidate their Representations}
\label{sec:coupling}
\begin{figure}[t]
\begin{minipage}[b]{\columnwidth}
  \centering
  \includegraphics[width=0.99\columnwidth]{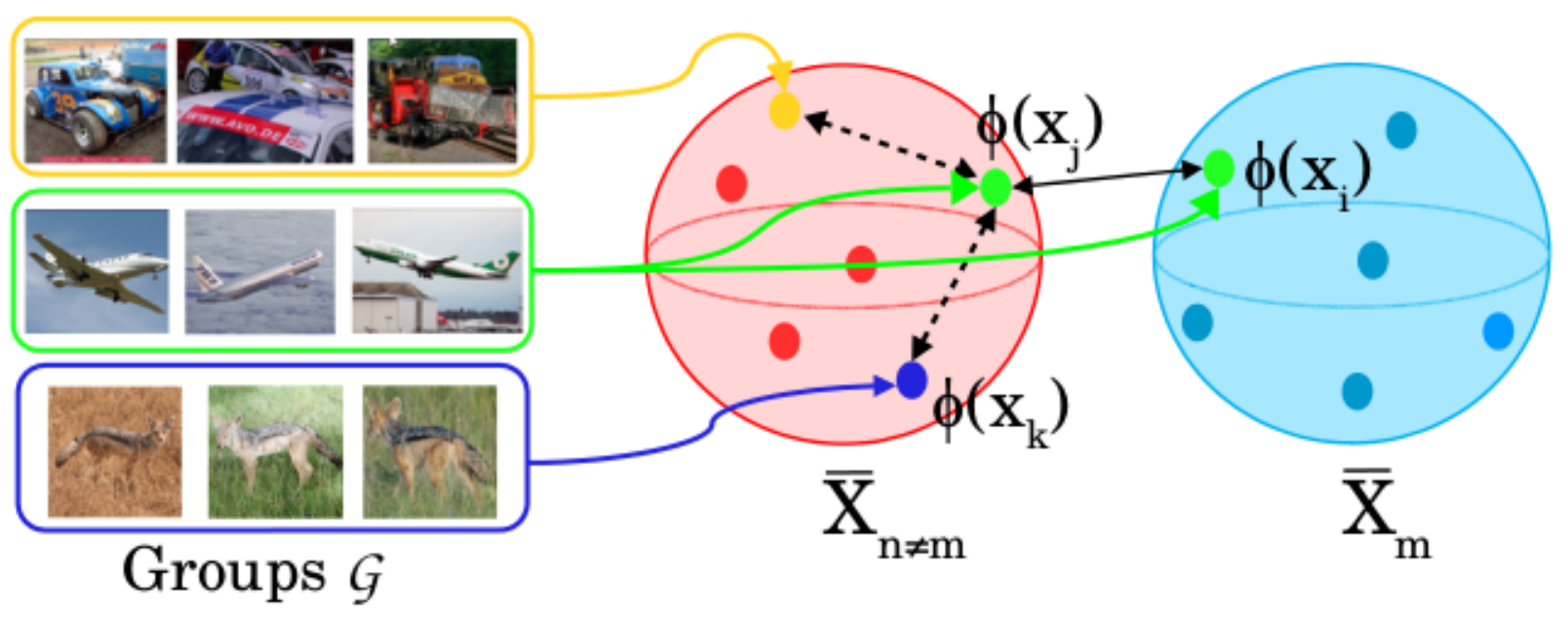}
\caption{
Triplet constraints for consolidating subproblems: A reliable relations between two subsets $\bar{X}_m$ and $\bar{X}_{n \neq m}$ couples them locally: solid black arrow between $x_i$ and $x_j$, which appear in the same group $G$ (green group). Further, we find $x_k \in \bar{X}_{n\neq m}$ which are reliably dissimilar to $x_j$ (dashed black lines). We now have a triplet $(x_i,x_j,x_k)$ relating the $x_k$, which was previously unknown to subset $\bar{X}_m$.
}
\label{fig:merge}
\end{minipage}
\end{figure}

We now have an ensemble of representations $\phi_k$, each representing a different subset of the data. The goal is now to consolidate their underlying learned relationships $\{d_{\phi_k}(\cdot,\cdot) \}_k$ into a global representation reflecting all of the data. 
Therefore, we look for reliable relations between different subsets to establish local links that allows to locally transfer relations from one subset into the other. Groups $G$ which are (partially) shared among subsets due to their overlap and distribution to different subsets by the partitioning act as anchors for such links as illustrated in Fig. \ref{fig:merge}. Using transitivity we thus find triplets $(x_i, x_j, x_k)$ across subsets which allows to transfer information about previously unknown relations from one local representation to another.
\\
Let $x_i$ be a member of a subset $\bar{X}_m$, but not of $\bar{X}_n$, i.e. $\bar{X}_m \ni x_i \notin \bar{X}_n$. Similarly let $\bar{X}_n \ni x_j \notin \bar{X}_m$ and $\bar{X}_n \ni x_k \notin \bar{X}_m$. Assume $\exists G: x_i, x_j \in G$, thus providing a reliable similarity between $x_i, x_j$ and consequently between both subsets. Similarly, we assume $x_j,x_k$ to have a reliable dissimilarity. Using transitivity the triplet $(x_i, x_j, x_k)$ thus implies an ordering constraint under the representation $\phi_m$, i.e., we want $d_{\phi_m}(x_i, x_j) \overset{!}{<} d_{\phi_m}(x_i, x_k)$. Note that these are additional relations imputed from $\bar{X}_{n \neq m}$, which were previously not present in $\bar{X}_m$. Let $\mathcal{T}_m$ be the set of all such triplets deduced by transitivity between $\bar{X}_m$ and $\bar{X}_{n \neq m}$. We now incorporate this additional information by refining $\phi_m$ solving the triplet ranking problem \cite{wu2017sampling}

\begin{equation}
\mathcal{L}_{\text{transfer}}^m = \sum_{(x_i, x_j, x_k) \in \mathcal{T}_m} \Big[d_m(x_i, x_j) 
- d_m(x_i, x_k) + \gamma \Big]_+
\end{equation}

\noindent
The parameter $\gamma$ controls the margin between $x_j$ and $x_k$ with respect to $x_i$. Note that the relations between $\bar{X}_m$ and the other subsets are only sparse. Thus the potentially large computational complexity of the triplet ranking loss does not dominate the complexity of the overall approach. \\
However, optimizing this objective alone would ignore and potentially forget about the dense relationships in $\bar{X}_m$ exploited by $\mathcal{L}_{\text{local}}^m$. Thus we combine both objectives,

\begin{equation}
\mathcal{L}_{\text{refine}}^m = \mathcal{L}_{\text{local}}^m + \mathcal{L}_{\text{transfer}}^m .
\label{eq:refine}
\end{equation}

Optimizing (\ref{eq:refine}) retains the constraints from $\bar{X}_{m}$ while incorporating couplings to other subsets $\bar{X}_{n \neq m}$ to improve $\phi_m$. Due to the additional inter-set relations, $\phi_m$ effectively now covers more of $X$ than before the refinement.
\\
In the last training iteration, one representation $\phi \in \{\phi_k\}_k$ becomes the final global representation. We conducted experiments using different aggregation strategies (averaging, random selection, etc.) and observed that after the final iteration all $\phi_k$ capture the dataset $X$ nearly equally well, allowing to randomly select one.

\begin{algorithm}[h!]
\small
\SetKwFunction{ExtractCompactGroups}{ExtractCompactGroups}
\SetKwFunction{PartitionGroups}{PartitionGroups}
\SetKwFunction{SampleTriplets}{SampleTriplets}
\SetKwFunction{SampleRandomGroups}{SampleRandomGroups}
\SetKwFunction{BuildMaximalReliableGroup}{BuildMaximalReliableGroup}
\SetKwFunction{LocalReassignment}{LocalReassignment}
\SetKwFunction{RegressTargets}{RegressTargets}
\SetKwFunction{RegressTargetsAndRefine}{RegressTargetsAndRefine}
\SetKwInOut{Input}{Input}
\SetKwInOut{Output}{Output}

\Input{$\mathcal{X}$: Set of training images}
\Parameter{
$T$: Number of training iterations; $K$: Number of subsets $\bar{X}_k$ \\
}
\Output{$\phi$: global representation}

\BlankLine
\BlankLine
\emph{\textbf{Train initial representation $\phi_{\text{init}}$:}}

$\varphi_i^{\text{init}} \sim$ uniform distribution on unit hypersphere; $\forall x_i \in \mathcal{X}$ \\
$\phi_{\text{init}}$ $\leftarrow$ $\min$ $\mathcal{L}_{\text{local}}$ \tcp{Eq.2}
$\phi$ $\leftarrow$ $\phi_{\text{init}}$ \tcp{initialize $\phi$}
\BlankLine
\BlankLine
\emph{\textbf{Iterative learning:}}

\For{$iter\leftarrow 1$ \KwTo $T$}{
    \emph{Extract reliable image relations} \\

    $G^h_{\text{rand}} \leftarrow$ \SampleRandomGroups{$\mathcal{X}, \phi$} \tcp{different sizes $h$}
    $\mathcal{G} = \{\}$ \\
    \For{$i\leftarrow 1$ \KwTo $|X|$}{
       $G \leftarrow$ \BuildMaximalReliableGroup{$x_i,G^h_{\text{rand}}$} \\
       $\mathcal{G} \leftarrow \mathcal{G} \cup G$
    }
    
    $\{\bar{X}_k\}_{k=1}^K$ $\leftarrow$ \PartitionGroups{$\mathcal{G}, K$} \tcp{Sec.3.4}
    
    \BlankLine
    \BlankLine
    \emph{Train local representations $\phi_k$} \tcp{Sec. 3.5.2}
    \For{$k\leftarrow 1$ \KwTo $K$}{
        $\mu_l^k \sim$ uniform distribution on unit hypersphere; $\forall G \in \bar{X}_k$ \\
        $\varphi_t^k \sim \mathcal{N}(\mu_l^k, \Sigma)$; $\forall x_i \in G$ \tcp{gaussian around $\mu_l^k$}

        \While{not converged}{
            $\pi \leftarrow$ \LocalReassignment{$\phi_k, \bar{X}_k, \varphi_t^k$} \\
            $\theta_k \leftarrow$ \RegressTargets{$\bar{X}_k, \varphi_{\pi(i)}^k$} \\
        }
    }
    
    \BlankLine
    \BlankLine
    \emph{Refine local representations $\phi_k$} \tcp{Sec. 3.5.3}
    \For{$k\leftarrow 1$ \KwTo $K$}{
        $\mathcal{T}_k \leftarrow $ \SampleTriplets{$\{\bar{X}_s\}_{s=1}^K$} \\
        $\mu_l^k \sim$ uniform distribution on unit hypersphere; $\forall G \in \bar{X}_k$ \\
        $\varphi_t^k \sim \mathcal{N}(\mu_l^k, \Sigma)$; $\forall x_i \in G$ \tcp{gaussian around $\mu_l^k$}
       \While{not converged}{
            $\pi \leftarrow$ \LocalReassignment{$\phi_k, \bar{X}_k, \varphi_t^k$} \\
            $\theta_k \leftarrow$ \RegressTargetsAndRefine{$\bar{X}_k, \varphi_{\pi(i)}^k, \mathcal{T}_k$} \\
       }
    }

    \BlankLine
    \BlankLine
    \emph{Update global $\phi$ by randomly choosing $k^* \in \{1,\dots,K\}$} \\
    
    $\phi$ $\leftarrow$ $\phi_{k^*}$
}
\caption{Unsupervised Representation Learning by Extracting Reliable Image Relations.}
\label{alg:training}
\end{algorithm}

\paragraph{Initialization:} As our overall iterative approach starts from random network initialization, initially we have no representation $\phi$ provided to extract reliable (dis)similarities for the grouping process in Sect. \ref{sec:reliableGroups}. Therefore, we train the first iteration by optimizing only the problem $\mathcal{L}_{\text{local}}$ based on the whole training set and randomly sample individual target points for each image, yielding our first representation $\phi_{\text{init}}$. The iterative approach then gradually learns stronger representations $\phi(\cdot)$ from iteration to iteration by capturing more and more reliable relationships in our dataset.

\subsubsection{Pseudo-Code for summary}
For additional clarity we now present a pseudo-code overview for our iterative approach (cf. Algorithm \ref{alg:training}).


\section{Experiments}
Following we evaluate the performance of our model on large scale datasets and the usability of our learned representation for the tasks of classification, detection and segmentation. Further, we present ablation and analysis experiments providing insights into our iterative learning process.

\begin{table}[t]
\centering
    \begin{tabular}{c|l|c}
    \toprule
    \multicolumn{1}{}{} & \multicolumn{1}{l|}{Method} &  Acc@1 \\ 
    \midrule 
    \hline
     \multirow{6}{*}{\rotatebox[origin=c]{90}{rgb input}} & Supervised ~\cite{pmlr-v70-bojanowski17a}  &  59.7  \\
     & Random~\cite{norooziECCV16}    &  12.0 \\ 
     \cline{2-3}  \cline{2-3}
     & Colorization~\cite{Colorful2016}   & 35.2 \\
     & Jigsaw Puzzles~\cite{norooziECCV16}  &38.1 \\
     & BiGAN~\cite{donahue2016bigan}   &32.2 \\ 
     &  RotNet~\cite{gidaris2018unsupervised}    & 43.8 \\ 
     \hline
     \multirow{9}{*}{\rotatebox[origin=c]{90}{gradient (sobel) input}} & Supervised~\cite{pmlr-v70-bojanowski17a} & 57.4 \\  \cline{2-3}
     & NAT~\cite{pmlr-v70-bojanowski17a}   & 36.0 \\
     & Deep Cluster~\cite{caron2018deep} & 44.0 \\
     & Ours (Initialization)   & 30.0 \\
     & Ours (Round 1)   & 39.1 \\
     & Ours (Round 2)   & 44.6 \\
     & Ours (Round 3)   & 45.8 \\
     & Ours (Round 4)    & \textbf{46.0} \\ \cline{2-3}
     & Ours (mean$\pm$std)  & 45.8$\pm$0.3 \\
    \bottomrule 
  \end{tabular}
  \caption{Comparison of our method to other state of the art unsupervised learning approaches on the ImageNet dataset. We report classification accuracy ($Acc@1$).}
\label{tab:imagenet_nonlinear}
\end{table}
  
\begin{table}[t]
\centering
  \begin{tabular}{@{}lcccc@{}}
    \toprule
    \multirow{2}{*}{Method} & Class & Det & Seg\\
&   (\%mAP) & (\%mAP) & (\%mIoU) \\
    \hline
    Trained layers  & all & all & all\\
    \hline \hline
    Supervised  & 79.9 & 56.8 & 48.0 \\
    Random &57.0 & 44.5 & 30.1\\
    \hline \hline
    Colorization~\cite{Colorful2016} &  65.6 & 46.9 & 35.6\\
	BIGAN~\cite{donahue2016bigan} & 60.1 & 46.9 & 34.9\\
    Jigsaw Puzzle~\cite{norooziECCV16} & 67.6 &53.2 & 37.6\\  
    NAT~\cite{pmlr-v70-bojanowski17a} & 65.3 & 49.4 & -\\  
    Split-Brain~\cite{zhang2017split} & 67.1 & 46.7 & 36.0\\   
    Counting~\cite{noroozi2017representation} & 67.7 & 51.4 & 36.6\\
    RotationNet~\cite{gidaris2018unsupervised} & 73.0 & 54.4 & 39.1\\
    DeepCluster~\cite{caron2018deep}
& 73.7 & 55.4 & \textbf{45.1} \\
\hline
    \hline
    Ours & \textbf{74.2} & \textbf{55.6} & 44.6\\ \hline
    Ours (mean$\pm$std) & 74.1$\pm$0.3 & 55.5$\pm$0.2 & 44.5$\pm$0.3\\ 
    \bottomrule    
  \end{tabular}
\caption{Comparing our model to state-of-the-art unsupervised approaches on PASCAL VOC 2007 classification, detection, and segmentation (measured in mean average precision and mean intersection over union.}
\label{tab:pascal_nonlinear}
\end{table}

\subsection{Implementation Details and Benchmarking}
We now evaluate our learned image representation on the ImageNet and PASCAL VOC dataset. We test on different vision tasks, such as image classification, objection detection, and semantic segmentation and compare our approach against state-of-the-art unsupervised methods. As preprocessing we convert our images to gradient images (obtained using a sobel filter) to avoid trivial solutions based on color, thus following the protocol of recent approaches \cite{pmlr-v70-bojanowski17a, caron2018deep}. These also conducted experiments on supervised ImageNet classification and concluded that gradient images yield similar performance in comparison to RGB inputs, thus indicating a fair comparison. If not stated otherwise, for all experiments the number of subsets $\bar{X}_k$ used for training is fixed to $K=5$. In the grouping stage we set $p$ to be the $3\%$ percentile. We set the dimensionality of $\phi_k$ to $D=2048$ and choose the parameters $\lambda_1, \lambda_2, \Sigma$ using cross-validation on the training set. The margin parameter $\gamma$ is set to $0.2$ as suggested in various works \cite{wu2017sampling}. We train our model using stochastic gradient descent using an initial learning rate of $0.01$ and momentum of $0.9$. Building groups can be efficiently performed using the FAISS \cite{faiss} library for fast nearest-neighbor retrieval with GPU support, thus leading to no significant computational overhead.

\paragraph{ImageNet:} We evaluate the ability of our model to capture differences between objects and its ability to scale to large image collections on the ImageNet dataset \cite{imagenet}. This dataset is composed of 1.2M images distributed over 1,000 categories including subtle category boundaries (such as dog breeds). For fair comparison with other methods we use the AlexNet \cite{alexnet} architecture with batch normalization. Our evaluation follows the standard protocol for unsupervised ImageNet pretraining: First, performing our unsupervised training on a randomly initialized network without using any labels. Afterwards, the convolutional layers are fixed while the last layers are randomly reinitialized and trained using ImageNet labels. Table \ref{tab:imagenet_nonlinear} (left) compares our results with other state-of-the-art unsupervised approaches. Our method converges to its final performance of $46.0\%$ after $4$ training iterations (excluding initialization). Hence, we are significantly improving upon all other unsupervised approaches including DeepCluster~\cite{caron2018deep} by $2\%$, which is trained on all training data at the cost of also incorporating unreliable noisy information. In contrast, our model successfully leverages more and more reliable image relations over the iterations, thus alleviating the issue of noise. Additionally we report mean and standard deviation of our approach (Ours (mean$\pm$std)) over 5 runs. Note that all of the other methods are only reporting their best run.

\paragraph{PASCAL VOC:} We now illustrate the generalization capability of our learned representation on different transfer learning tasks. We utilize our representation trained on ImageNet without labels (using the same architecture as above) and fine tune it on the PASCAL VOC 2007 \cite{pascal-voc-2007} classification, detection, and segmentation tasks (VOC 2012). For transfer learning we use the framework of Kr\"ahenb\"uhl et al.~\cite{krahenbuhl2015data} for classification experiments, the Fast R--CNN~\cite{girshickICCV15fastrcnn} framework for object detection and the method of Long et al.~\cite{fcn_semseg} for semantic segmentation. Our results are summarized in Table \ref{tab:imagenet_nonlinear} (right). On all transfer learning tasks, i.e. classification, detection and segmentation, our approach achieves comparable results to the unsupervised state-of-the-art which further demonstrates the expressive power of our representation. Additionally we report mean and standard deviation of our approach (Ours (mean$\pm$std)) over 5 runs. Note that all of the other methods are only reporting their best run.

\begin{table}[t]
\centering
\begin{tabular}{l|c}
\toprule
Method & Acc (\%)
\\ \hline   \hline
Target coding~\cite{yang2015targetcode} (supervised) & $73.2$  \\ \hline
Wang et al. \cite{csvddnet} & $68.2$ \\
CliqueCNN~\cite{cliqueCNN} & $69.3$ \\
Exemplar CNN~\cite{exemplarCNN} & $75.4$ \\
Discr. Attr.~\cite{Huang2016UnsupervisedLO} & $76.8$ \\
Chang et al. \cite{chang2018} & $74.6$ \\
\hline \hline
Ours & 75.3  \\ \bottomrule
\end{tabular}
\caption{Comparison to other approaches based on average classification accuracy on STL-10, using the unlabeled split for training.}
\label{tab:stl10_performance}
\end{table}

\subsection{Analysis and Ablation}
We now show ablation experiments and analyses to evaluate the individual parts of our approach and to provide insights into the iterative learning procedure.

\subsubsection{Ablation studies}
Ablation studies are conducted on the STL-10 dataset and summarized in Tab. \ref{tab:stl10_performance} (Left) and (Right).

\paragraph{STL-10 Performance:}
We contrast our approach to state-of-the-art methods on STL. For a fair comparison, we use the same network architecture as \cite{exemplarCNN} and train our model on the unlabeled split of the dataset. For this experiment we set $D=128$. Tab.~\ref{tab:stl10_performance} (Left) shows that our proposed approach achieves competitive performance to unsupervised state-of-the-art approaches, ExemplarCNN~\cite{exemplarCNN}, Discriminative Attributes~\cite{Huang2016UnsupervisedLO} and Chang et al.~\cite{chang2018}. However, in contrast to ExemplarCNN \cite{exemplarCNN} and Discriminative Attributes \cite{Huang2016UnsupervisedLO} whose learning procedures rely on dense, costly (instance-level-)exemplar classification tasks, our approach leverages only a sparse set of reliable constraints and thus scales to large datasets as shown on ImageNet. Chang et al.~\cite{chang2018} leverages individual images in combination with strong augmentations, thus neglecting valuable information from direct relations between training samples. Further, they are operating on a more powerful network architecture leading to unfair comparison in their favor. Note that our approach outperforms CliqueCNN~\cite{cliqueCNN} by $6\%$ which also learns from relations inferred by a grouping of images, however, without considering their reliability. This strongly indicates that the concept of reliability actually helps to reduce the amount of noise introduced into the learning process.

\begin{table}[t]
\centering
\begin{tabular}{l|c}
\toprule
Method & Acc. (\%)
\\ \hline \hline
Ours (Initialization) & $67.5$ \\
Ours (Round 1) & $71.2$ \\
Ours (Round 2) & $72.8$  \\
Ours (Round 3) &  $75.2$\\
Ours (Round 4) &  $75.3$ \\ \hline \hline

No decomposition into $\bar{X}_k$ (Round 1) & $69.1$ \\
Triplets Only (Round 1) & $62.6$ \\ \bottomrule
\multicolumn{2}{c}{} \\
\end{tabular}
\caption{Average classification accuracy on STL-10 for ablations of our approach.}
\label{tab:stl10_ablations}
\end{table}

\paragraph{Divide-and-Conquer:} The performance of our representation increases with each iteration and improves over our initial representation by $7.8\%$, cf. Tab. \ref{tab:stl10_performance} (Right). Now, to evaluate the effect of incorporating reliable dissimilarities, we train our model using only a single target space (No decomposition) for all groups for one iteration after initialization. As a result there are only reliable relations within the $G\in\mathcal{G}$, but lots of unreliable relationships between them. Due to the former, performance slightly increases over the initialization by $1.6\%$. The latter explains the $2.1\%$ lower performance compared to our divide-and-conquer strategy. This highlights the importance of modelling reliable dissimilarities when learning visual representations.

\paragraph{Full model vs. triplet learning:} In this ablation experiment (Tab.\ref{tab:stl10_performance} (Right) Triplets Only) we use our extracted reliable (dis-)similarity relationships to mine triplet-constraints as input into a standard triplet-loss framework \cite{wu2017sampling}. For a sample that serves as triplet anchor, reliable similarity relations act as positive constraints and reliable dissimilarity relations act as negative constraints. The aim of this experiment is to contrast our learning objective, Eq. (4), against popular ranking loss approaches. The massive drop of $8.6\%$ in performance can be explained by the dependence of such frameworks on hard-negative mining strategies. Since reliable constraints are only based on high similarities and dissimilarities, the ranking framework obviously has no access to hard constraints, which are very difficult to find reliably \textit{without} supervision \cite{wu2017sampling} or strong pretraining \cite{fuzzy_dml}. Note that we are using triplet constraints only for transferring already learned information to refine our representations $\phi_k$ (Sec. \ref{sec:coupling}) rather than using them as the driving overall learning signal.

\paragraph{Number of subsets:}
To evaluate the sensitivity of our approach with respect to the number of subsets $\bar{X}_k$ used for training, we train multiple models using different values of $K$. Tab. \ref{tab:stl10_ablation_numSpheres} illustrates, that training performance saturates for $K > 10$ subsets. This indicates that further partitioning of the data and thus further maximizing the dissimilarity between groups in $\bar{X}_k$ has no effect and has been achieved to a sufficient degree.

\subsubsection{Analysis of the model:} We now further analyze the effect of iterative training based on the model used for the ImageNet benchmark. The following experiments highlight the ability of our model to turn reliable (dis-)similarities into increasingly correct sample relations while simultaneously harnessing more and more data.

\begin{figure}[t]
  \centering
  \includegraphics[width=0.99\columnwidth]{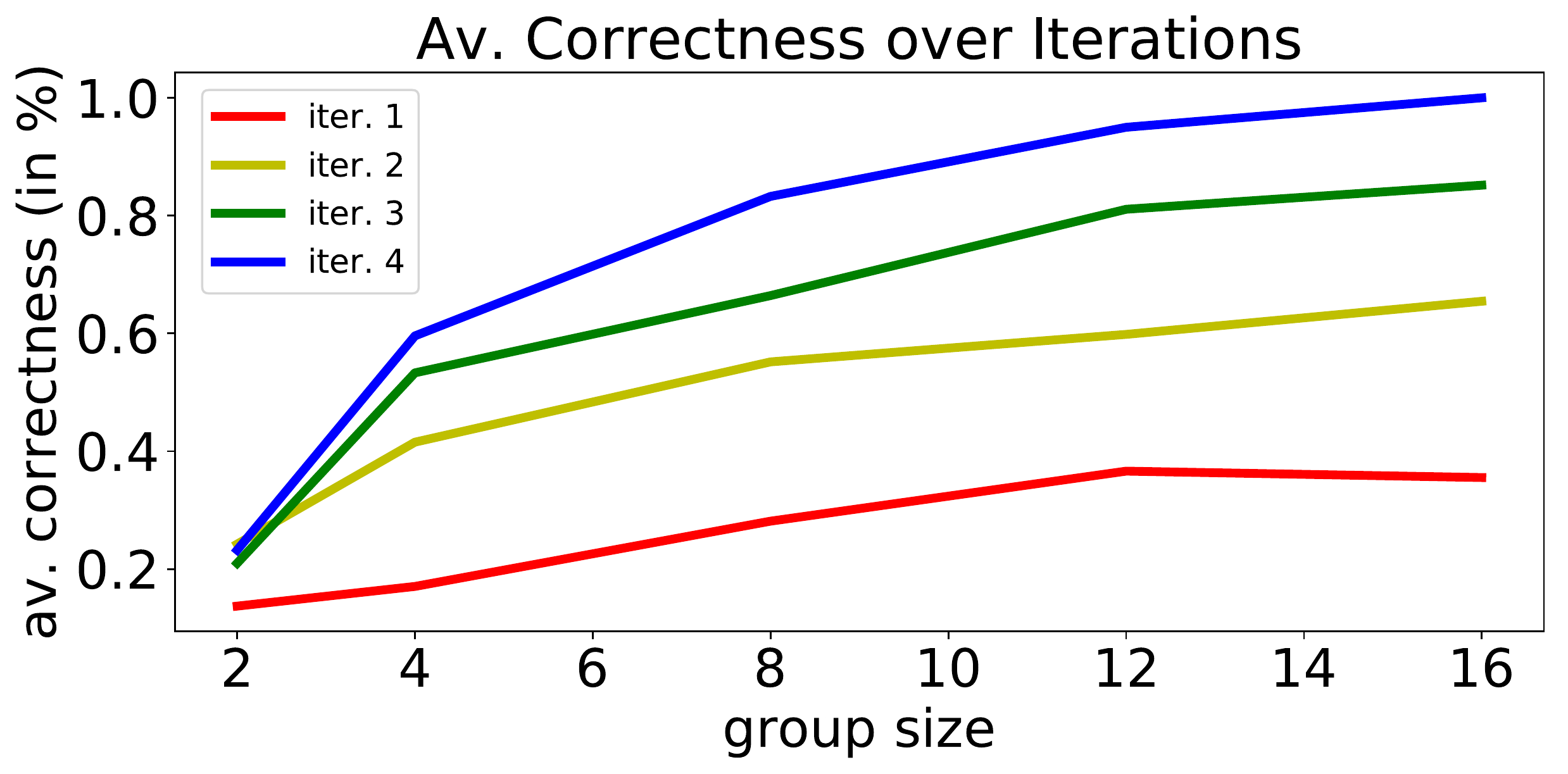}
\caption{Average correctness for groups $G$ of different size and in different training iterations. Average correctness is the fraction of members with the same ground-truth ImageNet class label (not used for training).}
\label{fig:correct_over_iterations}
\end{figure}

\paragraph{Correctness of image relations in groups $G$:}
In Sec. \ref{sec:reliableGroups} and Sec. \ref{sec:partitioning} we gather reliable relations driving the learning of our representation. Fig. \ref{fig:purity_over_groups} (a) illustrates that our procedure of extracting groups $G$ for reliable relations increases the probability of finding correct relationships. Moreover, successive iterations of training improve the learned image relations. Fig. \ref{fig:correct_over_iterations} confirms this by measuring how many relations within a group disagree. One can see that in consecutive iterations, our groups $G$ exhibit more and more correct relationships. This proves that our model is able to extend the reliable relations learned from a group $G$ to relations which so far could not be identified as reliable. Consequently the performance of our model increases.

\begin{table}[t]
\centering
\begin{tabular}{l|l}
\toprule
\multicolumn{1}{c}{Number of Subsets $\bar{X}_k$} &\multicolumn{1}{c}{Acc. (\%)}
\\ \hline \hline \noalign{\vskip 1mm}
no partitioning & $69.1$ \\
2 subsets & $69.5$ \\
5 subsets & $70.8$ \\
10 subsets & $71.2$ \\
20 subsets & $71.0$  \\
\bottomrule
\end{tabular}
\caption{Average classification accuracy on STL-10 dataset as number of subsets $K$ is varied. Results are for one iteration of training after initialization.}
\label{tab:stl10_ablation_numSpheres}
\end{table}

\begin{figure}[t!]
  \centering
  \includegraphics[width=0.99\columnwidth]{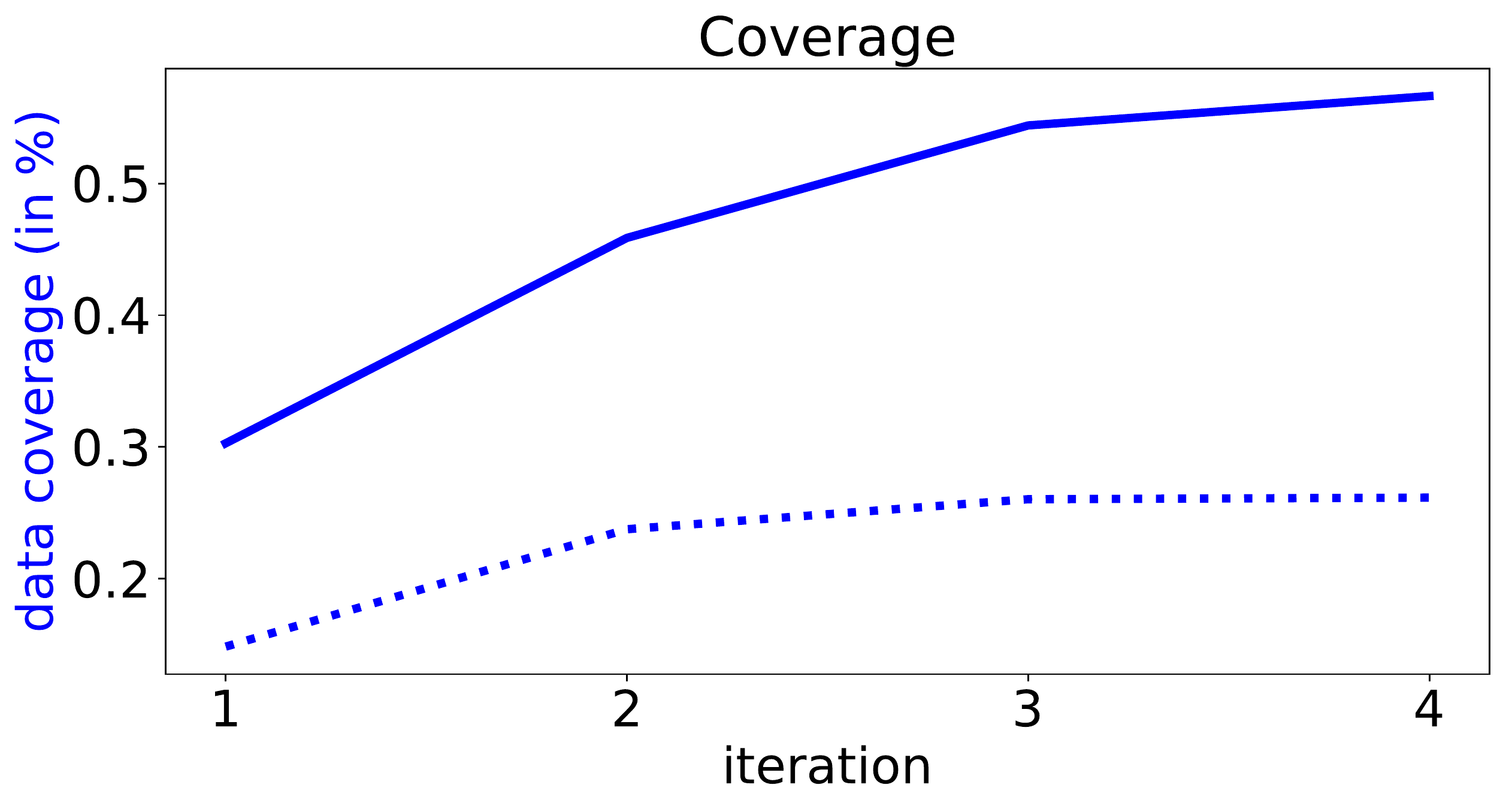}
\caption{Fraction of training data covered by subsets $\bar{X}_k$ in successive training iterations. Solid: overall data coverage. Dashed: mean data coverage per subset.}
\label{fig:overlap_coverage}
\end{figure}

\paragraph{Data coverage:}
In our proposed method we deliberately explore an orthogonal approach to simply using all training samples by reducing the noise introduced into the training process using only reliable relations for learning. Consequently there is a trade-off between exploring more data and introducing more noise due to unreliable relations. Fig. \ref{fig:overlap_coverage} shows the amount of training data covered in each training iteration. Overall and per subset, our grouping process covers more and more data samples, since representations improve and more relations become reliable. Between iteration $1$ and iteration $4$ the amount of data available for training $\phi$ is almost doubled to $60\%$. Moreover, Fig. \ref{fig:correct_over_iterations} proves that the additional data is meaningful and not just noise, since overall data quality is improving simultaneously. This demonstrates that our model not only reinforces already available reliable relations but is able to generalize its representation to previously unused data, i.e., discovering new reliable relationships. Further, the result that we achieve state-of-the-art performance using 60$\%$ of the available data proves $\emph(i)$ there exists an alternative way to using all but therefore noisy relations between training samples, $\emph(ii)$ we are able to successfully find and exploit reliable samples and $\emph{(iii)}$ the idea of considering data reliability for unsupervised representation learning is not fully solved, thus opening new promising directions for future research.

\begin{figure*}[h!]
  \centering
  \includegraphics[width=18cm]{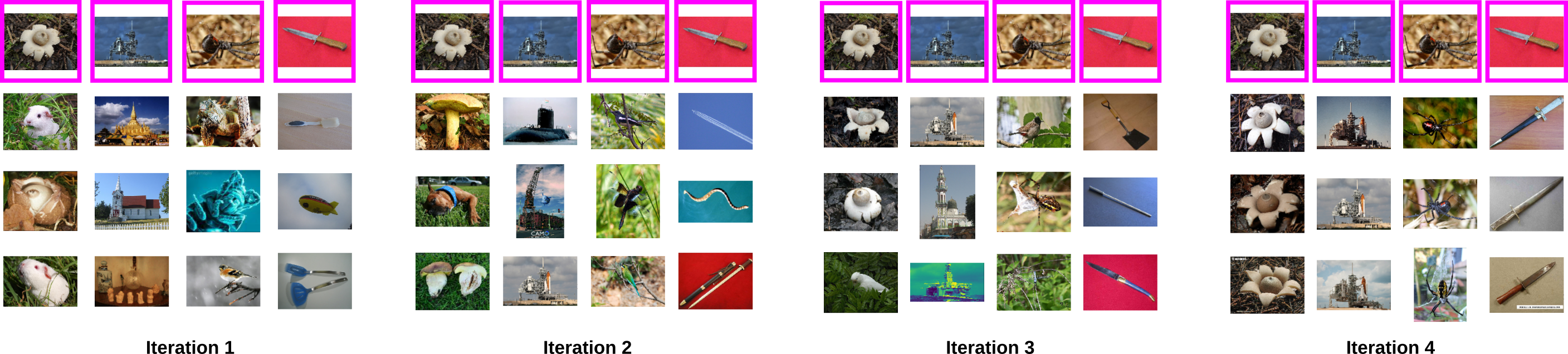}
\caption{Groups of size $4$ for fixed seed image (pink) over the iterations while training the ImageNet model. Each column represents a group $G$.}
\label{fig:group_evo_4}
\end{figure*}

\begin{figure*}[h!]
  \centering
  \includegraphics[width=18cm]{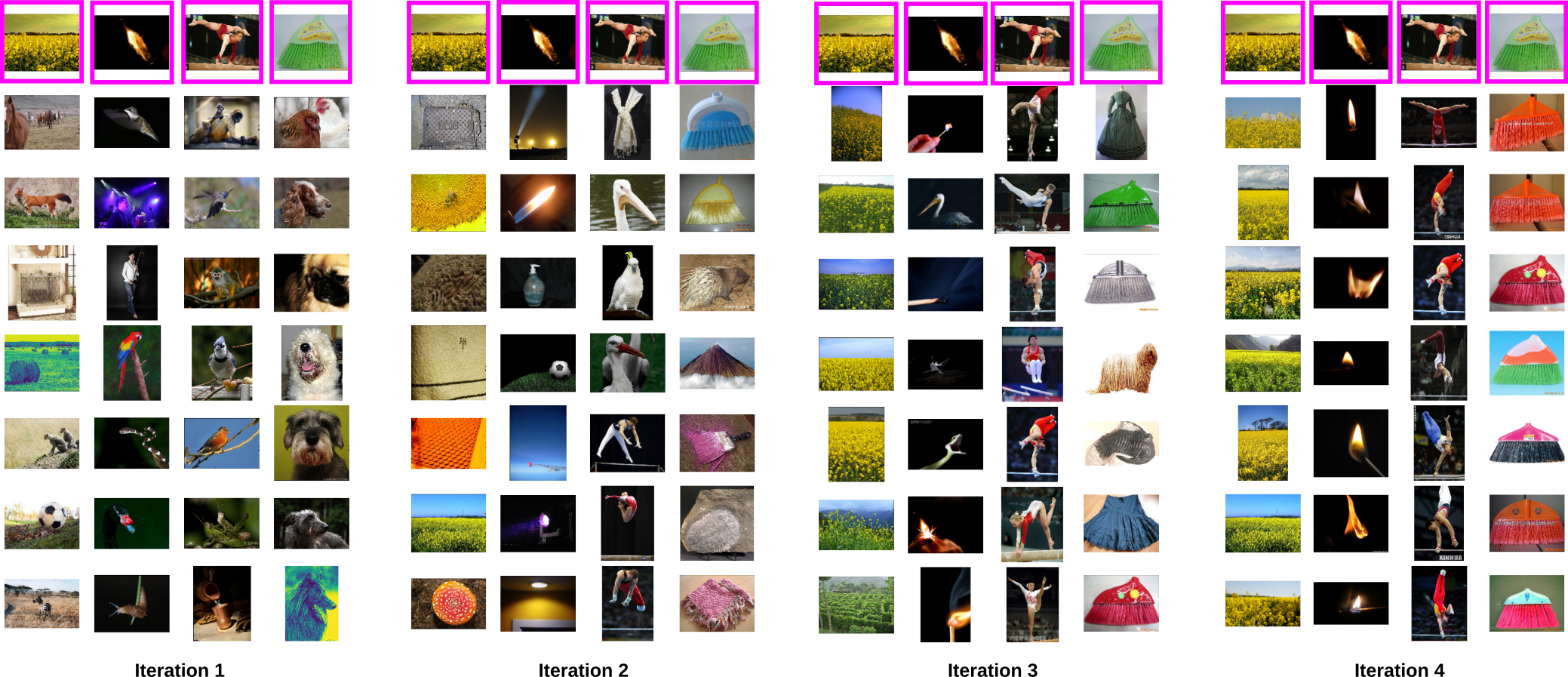}
\caption{Groups of size $8$ for fixed seed images (pink) over the iterations while training the ImageNet model. Each column represents a group $G$.}
\label{fig:group_evo_8}
\end{figure*}

\paragraph{Fixing the seed of groups $G$}
In Fig. \ref{fig:group_evo_4} and Fig. \ref{fig:group_evo_8} we present examples for groups of size $4$ and $8$ while training on the ImageNet dataset. To allow for a detailed comparison, we fix the seed elements $x_i$ and show how the constituents of a group change over the iterations. We observe that over the iterations the constituents of each group share more and more meaningful visual features. At the beginning the representation focuses on coarse visual commonalities like rough shape and scene. At the end relationships between constituents are dominated by true (intra-)class-specific features and similar pose. 

\begin{figure*}[h!]
  \centering
  \includegraphics[width=14cm]{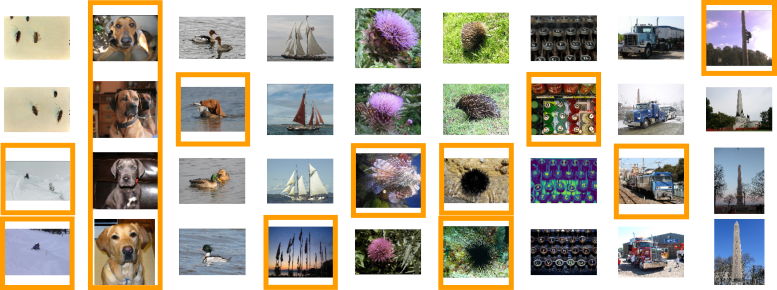}
\caption{Examples of groups with incorrect constituent relationships (i.e. different labels) taken from the last iteration while training on ImageNet. Orange highlights the source of disagreement. Each column represents a group $G$.}
\label{fig:group_fail}
\end{figure*}

\paragraph{Typical sources of incorrectness}
To explain common sources of incorrect relationships within groups, i.e., group constituents having different labels, we show typical examples in Fig. \ref{fig:group_fail}. As one can see, disagreements often arise due to subtle differences between the constituents' classes (such as different dog breeds, hedgehogs vs. sea urchins, etc.) and misleading scene settings (such as a buildup of flags imitating a ship's shape, a dog's head above the surface while swimming looking like a duck, etc.).

\paragraph{Analysis of computational cost}
As deriving exact computational complexities is often difficult, it is common practice for deep learning based methods to compare their computational complexities based on their training times. For our model learning one representation $\phi_k$ (including the refinement step) for a subset $\bar{X}_k$ takes approx. 16h on a Titan X (Pascal), resulting in 14 days of training in total for the Imagenet dataset. Thus, our overall training time is comparable to the current state-of-the-art approach DeepCluster \cite{caron2018deep} (12 days). Further, Tab. \ref{tab:training_time} compares the computational cost of our method with recent unsupervised representation learning approaches in relation to their performance. As we observe, peak performances on Imagenet are computationally costly. Further, we observe that RotNet~\cite{gidaris2018unsupervised} offers the best trade-off between training efficiency and performance.

\section{Conclusion and Future Work:}
In our manuscript we have presented a novel iterative approach for unsupervised learning of visual feature representations. Experimental evaluation shows that our method yields a representation of competitive or even superior performance on the tasks of unsupervised classification and transfer learning compared to the state-of-the-art. We propose a novel technique for finding reliable relations (i.e. dis-/similarities) between training images which are likely to agree with the ground-truth. This reduces the amount of noise due to erroneous relations introduced into training, which is typically an issue when all possible training data relations are directly used. As using all relations is typically common practice in the vast majority of unsupervised learning literature, our work offers a new direction of future research directions: Instead of solely looking for more powerful surrogate tasks to compensate for missing supervision, we also address the question which training samples and relations can be trusted and which are likely to obstruct learning. This question naturally leads to a trade-off between covering all available training data and exploiting only reliable relations for learning. Hence, future work should focus on further improving the estimation of reliability to increase data coverage while maintaining a high reliability for efficient learning. Further, as the concept of reliable relations is of general value and can be decoupled from the learning itself, we investigate its applicability in unsupervised learning in general.
\\
We presented a technique for identifying reliable relations resulting in a set of local sub-problems. For each subproblem the data samples are either reliably similar or reliably dissimilar. Further, the learning process for each subproblem is formulated to preserve their reliable relations and approximate the actual distribution of distances of the training data. This stands in contrast to previous approaches whose feature space relies on data-independent prior assumptions which potentially disagree with the training data at hand. We then optimize each problem individually before using transitivity relations between them to efficiently merge their learned local representations into a single concerted representation. To increase the efficiency of our learning procedure, another line of future work addresses the trade-off between peak performance and computational cost. To this end, we investigate possible approximations of the introduced optimization and learning problem and analyze their implications on the learned representation.

\begin{table}[t]
\centering
\begin{tabular}{l|l|l}
\toprule
\multicolumn{1}{c}{Method} & \multicolumn{1}{c}{Running time} & \multicolumn{1}{c}{Acc@1}
\\ \hline \hline \noalign{\vskip 1mm}
Supervised & $3$ days (Titan X) & $59.7$ \\
\hline
NAT~\cite{pmlr-v70-bojanowski17a} & $1$ days (Titan X) & $36.0$\\
RotNet~\cite{gidaris2018unsupervised} & $2$ days (Titan X) & $43.8$ \\
Jigsaw Puzzle~\cite{norooziECCV16} & $3$ days (Titan X) & $38.1$ \\
BiGAN~\cite{donahue2016bigan} & $3$ days (Titan X) & $32.2$ \\ 
DeepCluster~\cite{caron2018deep} & $12$ days (Titan X) & $44.0$ \\
Ours & $14$ days (Titan X) & $46.0$ \\
\bottomrule
\end{tabular}
\caption{Training time vs. performance on the ImageNet dataset. Comparison of training times is based on the reported timings in the manuscripts of each method using a single NVIDIA Titan X (Pascal). Additionally, their performance for Acc@1 on ImageNet is reported.}
\label{tab:training_time}
\end{table}

\section*{Funding}
This work has been supported in part by DFG grant OM81/1-1 and a hardware donation from NVIDIA Corporation.

\section*{Author Biographies}
\textbf{Timo Milbich} received his masters degree in Scientific Computing
from Ruprecht-Karls-University, Heidelberg, in 2014. He
is currently a Ph.D. Candidate in Heidelberg Collaboratory for
Image Processing at Heidelberg University. His current research
interests include computer vision focusing on deep representation and metric learning. \\
\indent
\textbf{Omair Ghori} received his masters in Information and Communication Engineering from TU Darmstadt. He is currently a senior data scientist at AGT International and is also pursuing his PhD at the Heidelberg Collaboratory for Image Processing at Heidelberg University. His research interests include Action Recognition and deep representation learning. \\
\indent
\textbf{Ferran Diego} received his Masters Degree and PhD in Computer Science from Autonomous University of Barcelona. He was a postdoctoral researcher in the IAL group at Heidelberg University, Germany. He is currently a Research Scientist at Telefonica R\&D. His research interests include computer vision, machine learning, deep learning and neuroscience. \\
\indent
\textbf{Bj\"orn Ommer}  received a diploma in computer science from the University of Bonn, Germany and a Ph.D. from ETH Zurich. After holding a postdoctoral position at the University of California at Berkeley he has joint the Department of Mathematics and Computer Science at Heidelberg University as a professor where he is heading the computer vision group. His research interests include computer vision, machine learning, and cognitive science.



\bibliographystyle{elsarticle-harv} 
\bibliography{egbib}





\end{document}